\documentclass[11pt]{article}

\usepackage{acl}
\usepackage{booktabs}
\usepackage{multirow}

\usepackage{times}
\usepackage{latexsym}
\usepackage{amsmath}
\usepackage[table]{xcolor}

\definecolor{bestyellow}{RGB}{204, 225, 240} %
\definecolor{secondgray}{gray}{0.85}         %

\newcommand{\best}[1]{\cellcolor{bestyellow}{#1}}

\usepackage{listings}

\usepackage[T1]{fontenc}

\usepackage[utf8]{inputenc}

\usepackage{microtype}

\usepackage{inconsolata}

\usepackage{graphicx}
\usepackage{amssymb}

\title{EnterpriseBench: Benchmarking LLM Agents on Enterprise-Level Strategic Reasoning and Decision-Making}

 \author{%
    Min Yang$^{1,2}$, Yichen Pan$^{2,4}$, Jinghua Piao$^{2,3 ~\textbf{*}}$, Dandan Song$^{4}$, Yongshun Gong$^{1}$, \\
    \textbf{Yong Li}$^{2,3}$ \thanks{Jinghua Piao and Yong Li are co-corresponding authors.}\\ 
    $^{1}$Shandong University \quad                                             
    $^{2}$Zhongguancun Academy \quad                                             
    $^{3}$Tsinghua University \\                                                 
    $^{4}$Beijing Institute of Technology \\ 
    minyang@mail.sdu.edu.cn \quad   
    Pjh22@mail.tsinghua.edu.cn   \quad                          
    {3120230993@bit.edu.cn} \quad  \\   
    sdd@bit.edu.cn \quad
    ysgong@sdu.edu.cn \quad 
    liyong07@tsinghua.edu.cn
  }
  
\begin{document}
\maketitle

\raggedbottom
\begin{abstract}
LLM agents are increasingly expected to support enterprise workflows, where tasks often involve missing information, uncertainty, feedback, and long-term trade-offs. However, existing enterprise and financial benchmarks mainly test static capabilities such as information extraction, numerical calculation, domain knowledge, and financial QA, leaving interactive and long-horizon decision-making underexplored. To bridge this gap, we introduce \textbf{EnterpriseBench}, a benchmark that evaluates LLM agents across this spectrum, from static question answering to dynamic decision-making. Specifically,  EnterpriseBench reorganizes existing enterprise and financial QA datasets into a unified foundational suite annotated by capability and difficulty, and introduces three professional interactive settings: \textit{Consulting}, based on management-consulting-style business cases for client problem diagnosis through multi-turn information seeking; the \textit{Beer Game}, adapted from a classic supply-chain management simulation for inventory control under delayed feedback; and \textit{Enterprise Digital Twin}, a project-based business simulator for workforce, risk, and project planning. Experiments with nine agent methods under four backbone models show that current agents have not yet achieved stable, comprehensive, and cross-task reliability in enterprise scenarios. These results show that EnterpriseBench provides a practical benchmark for evaluating LLM agents in realistic enterprise strategic reasoning and decision-making. Our codes and benchmark implementation are publicly available\footnote{\url{https://github.com/sduyangmin/FirmBench}}.

\end{abstract}

\section{Introduction}







Recent advances in Large Language Models (LLMs) and LLM-based agents have substantially improved their reasoning and decision-making capabilities \cite{liu2024deepseek, yang2025qwen3, zhang2025agent, zhang2025aflow, yu2024fincon, li2025investorbench, xing2025designing}. Beyond isolated problem solving, recent studies have begun to explore LLM agents in more complex financial decision-making scenarios, such as investment and trading \cite{raptis2025agentic, xiao2024tradingagents, chen2025stockbench}. These scenarios differ from conventional fixed-input tasks: agents must often gather missing information, reason under uncertainty, act in dynamic environments, observe delayed feedback, and adjust their strategies over time. This raises an evaluation challenge: benchmarks need to assess not only static reasoning ability, but also decision-making through interaction with dynamic environments.

However, existing financial and enterprise benchmarks still mainly evaluate static, fixed-input capabilities. They typically focus on financial report understanding, table-based numerical reasoning, domain knowledge, information extraction, and financial QA \cite{chen2021finqa, zhang2025xfinbench, mohammadi2025evaluation, zhu2024benchmarking}. These tasks are important for assessing foundational enterprise reasoning, but they remain far from real enterprise decision-making scenarios, where agents must actively acquire missing information, respond to delayed feedback, and make long-horizon plans under uncertainty. As a result, current financial and enterprise benchmarks provide limited evidence about whether LLM agents can support interactive and strategic decision-making in realistic enterprise workflows.

To address this gap, we draw on realistic enterprise scenarios that involve missing information, uncertainty, feedback, and long-term trade-offs to construct \textbf{EnterpriseBench}, a benchmark for strategic reasoning and decision-making.  EnterpriseBench is designed around a two-layer evaluation principle. The first layer assesses foundational capabilities that a competent enterprise agent should possess before making decisions. These include extracting evidence from documents, performing numerical calculations, applying domain knowledge, and answering complex reasoning questions. To support this layer, we reorganize existing enterprise and financial QA datasets into a unified evaluation suite covering \textit{Information Extraction}, \textit{Numerical Calculation}, \textit{Domain Knowledge}, and \textit{Complex Reasoning}. The second layer evaluates whether agents can use these foundational capabilities in interactive enterprise decision-making scenarios. It includes three tasks constructed from professional case materials and established management simulations: Consulting, Beer Game, and Enterprise Digital Twin (EDT). For Consulting, we first collect raw cases from MBA consulting casebooks distilled from real consulting firm interviews. We then convert the curated materials into simulated case interviews, where an LLM acts as the interviewer and interacts with the tested agent based on the case materials. For Beer Game and EDT, we adapt historically grounded management simulations into MCP based testing interfaces: Beer Game supports supply chain ordering decisions, while EDT supports project planning decisions in a enterprise simulator. Together, these tasks provide standardized environments for evaluating agents through simulated interviews and tool based business simulations.

We conduct extensive experiments with nine representative LLM agent methods under four backbone models, DeepSeek-V3, GPT-4.1, DeepSeek-V4Pro and GLM-5.2.  
The results show that current agents have not yet achieved stable, comprehensive, and cross-task reliability in enterprise scenarios: strong QA performance does not reliably transfer to Consulting, Beer Game, or EDT. 
We also observe that the best-performing method varies across tasks and backbones, with no single method consistently performing best. Accordingly, we propose a proof-of-concept method that dynamically assembles suitable agent methods for different task instances, improving the overall QA score from 0.725 to 0.729 over the best fixed method.
To assess benchmark reliability, we further conduct three checks: a human audit of Consulting evaluation on 20 cases, prompt robustness tests for Consulting, and a human audit of difficulty annotation on 50 tasks. The Consulting score changes by at most 0.12 points across prompt variants, and the difficulty annotation aligns strongly with human ratings, with Pearson $r=0.96$. These results support the reliability of EnterpriseBench's evaluation protocol and annotation process.
Overall, these results show that EnterpriseBench is a challenging benchmark with empirical reliability support, offering a realistic way to assess the strengths and limitations of LLM agents in enterprise strategic reasoning and decision-making.

Our contributions are summarized as follows:
\begin{itemize}
    \item We introduce EnterpriseBench, a unified benchmark that evaluates LLM agents from foundational financial and enterprise QA to interactive enterprise decision-making.
    \item We introduce simulated professional interviews as a new form interactive benchmark for analysis-intensive tasks, where test agents must clarify objectives, acquire missing information, structure analyses, and communicate recommendations through sustained interaction.
    \item We systematically evaluate nine representative LLM agent methods under four backbone models, showing that EnterpriseBench offers a challenging and reliable platform for characterizing the strengths and limitations of LLM agents in realistic enterprise strategic reasoning and decision-making.
\end{itemize}

\section{EnterpriseBench}
\begin{figure*}[th]
    \centering
    \includegraphics[width=\linewidth]{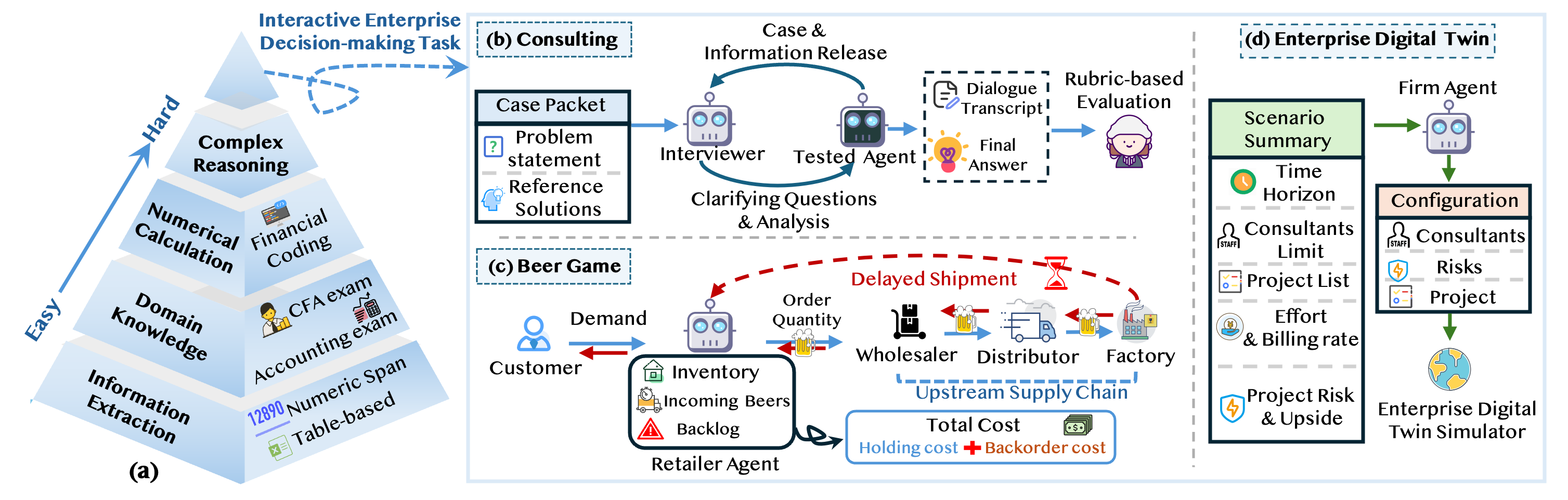}
    \caption{
Overview of EnterpriseBench. (a) EnterpriseBench organizes enterprise-oriented evaluation into a layered capability structure, progressing from information extraction, domain knowledge, and numerical calculation to complex reasoning. (b)--(d) The benchmark further introduces three interactive decision-making tasks: Consulting for hidden-information business problem solving, the Beer Game for delayed-feedback supply-chain control, and Enterprise Digital Twin for project-based enterprise operation.
}
    \label{fig:main_figure}
\end{figure*}

\subsection{Overview and Design Principles}

EnterpriseBench follows a two-layer evaluation principle: foundational enterprise reasoning and interactive enterprise decision-making. As shown in Fig.~\ref{fig:main_figure}, the foundational layer organizes enterprise-oriented evaluation into a layered capability structure, progressing from explicit evidence use and numerical computation to professional knowledge application and decision-oriented reasoning. The interactive layer then evaluates whether agents can apply these foundational capabilities when information is incomplete, feedback is delayed, and actions affect future outcomes. This layer introduces three tasks: Consulting for hidden-information business problem solving, Beer Game for delayed-feedback supply-chain control, and Enterprise Digital Twin for project-based enterprise operation.

\subsection{Foundational Enterprise Reasoning Tasks}
\label{sec:taxonomy}

The foundational component of EnterpriseBench selectively curates heterogeneous enterprise and financial QA datasets into a standardized evaluation suite. It is designed to assess the foundational capabilities required by enterprise agents across four levels, progressing from explicit evidence use to computation, professional knowledge, and decision-oriented reasoning.

\paragraph{Information Extraction.}
This category evaluates whether agents can locate, extract  explicit information from enterprise documents without external knowledge or calculation. Representative tasks include retrieving numerical facts from filings, identifying relevant spans in financial reports, and extracting values from tables or textual disclosures \cite{chen2021finqa, krumdick2024bizbench}. These tasks test whether agents can ground their answers in explicitly provided enterprise evidence.

\paragraph{Numerical Calculation.}
This category evaluates arithmetic operations, numerical reasoning, formula-based computation, and code-based calculation. Representative tasks require agents to compute changes, percentages, or formula outputs from financial tables and textual evidence, and in some cases to write executable code for financial QA \cite{krumdick2024bizbench, zhang2026agentic}. These tasks test whether agents can transform enterprise evidence into quantitative answers.

\paragraph{Domain Knowledge.}
This category evaluates specialized business and financial knowledge itself, including finance and accounting concepts, XBRL or US-GAAP tags, financial formulas, accounting standards, CFA-style knowledge questions, and business-related exam problems focused on conceptual understanding rather than decision-making \cite{krumdick2024bizbench, zhang2026agentic}. These tasks test whether agents understand the professional concepts, standards, and terminology needed for enterprise reasoning.

\paragraph{Complex Reasoning.}
This category evaluates judgment and decision-making under stated constraints. Representative tasks include scenario-based CFA questions and business-related exam problems in business ethics, microeconomics, and professional accounting that ask what action should be taken in a given situation \cite{krumdick2024bizbench}. These tasks test whether agents can make context-sensitive decisions rather than only recall knowledge, extract information, or perform calculations.
All samples are converted into a common input-output format and annotated with both a capability category and a 0--10 difficulty score. Difficulty is assigned based on context length, reasoning depth, ambiguity, label-space complexity, and required reasoning steps, and grouped into \textit{Easy} (0--3), \textit{Middle} (4--6), and \textit{Hard} (7--10). The annotations are initialized using an expert LLM with dedicated prompts and validated through human audit in Section~\ref{sec:validity}; the full prompt is provided in Appendix~\ref{sec:reclass_prompts}. Through this selective curation and annotation, EnterpriseBench evaluates whether agents can retrieve enterprise evidence, perform financial computation, apply professional knowledge, and integrate these capabilities for decision-oriented reasoning.

\subsection{Interactive Decision-Making Tasks}

To evaluate capabilities beyond static answer generation, EnterpriseBench introduces three interactive enterprise decision-making tasks. These tasks cover complementary forms of enterprise decision-making complexity: hidden information in business problem solving, delayed feedback in operational control, and long-horizon planning in enterprise simulation.

\paragraph{Consulting.}
The Consulting task is derived from MBA consulting casebooks and consulting-club preparation materials, many of which are based on real cases used to train candidates for consulting-firm interviews. Unlike standard QA tasks that emphasize answer correctness, these cases evaluate structured problem solving, strategic reasoning, information seeking, quantitative analysis, and communication under partial information.

As shown in Fig.~\ref{fig:main_figure}(b), we convert the curated case materials into simulated interviews, where an LLM interviewer interacts with the tested agent based on the case materials. We curate 423 cases, each containing a visible problem statement, hidden case facts, and a reference solution. The agent receives only the initial business problem and must acquire hidden information through relevant clarification questions, structure its analysis, perform quantitative reasoning when needed, and produce a final business recommendation.
After the interview, an LLM-based judge evaluates the full dialogue and final recommendation along four dimensions: structure, quantitative reasoning, business sense, and communication. The overall Consulting score is computed as the average of these dimensions. We further validate this LLM-based evaluation against human judgments in Section~\ref{sec:validity}.

\paragraph{Beer Game.}
The Beer Game is adapted from the classic Beer Distribution Game~\cite{Sterman1989} in the system-dynamics tradition, a long-standing management simulation for studying supply-chain coordination, delayed feedback, inventory oscillation, and the bullwhip effect. As shown in Fig.~\ref{fig:main_figure}(c), we use it to evaluate sequential operational decision-making under partial observability. The environment simulates a four-stage supply chain consisting of a retailer, wholesaler, distributor, and factory. The tested agent controls the retailer, observes only local state variables such as inventory, backlog, incoming shipments, and realized customer demand, and chooses an order quantity at each time step.

The challenge arises from delayed consequences and non-stationary dynamics. Under-ordering leads to backorder costs, while over-ordering increases inventory holding costs; other participants follow fixed equation-based policies unknown to the agent, and the environment includes shipment delays and a demand shift. We run each Beer Game episode for 25 time steps and evaluate performance by total accumulated cost, with lower cost indicating better decisions.

\paragraph{Enterprise Digital Twin.}
EDT is adapted from the transentis Enterprise Digital Twin project, which originated from transentis's 2023 Enterprise Digital Twin meetup series and models the company's professional-service business. As shown in Fig.~\ref{fig:main_figure}(d), EDT evaluates long-horizon enterprise operation in a project-based business simulator. Unlike the Beer Game, where the agent acts at every time step, EDT adopts a scenario-level decision protocol: the agent receives a template enterprise scenario describing the simulation horizon, workforce, candidate projects, workloads, revenue rates, risks, and default schedules, and outputs a complete enterprise strategy.

The strategy specifies workforce size, a global revenue risk level, and a project portfolio schedule. For each project, the agent decides whether to accept it and, if accepted, sets its start and deadline steps. The resulting configuration is materialized as a new EDT scenario and executed for a full episode, where the simulator computes revenue, expenses, utilization, profit margin, and accumulated earnings. EDT therefore requires agents to balance revenue opportunities against labor costs, capacity constraints, deadlines, and risk-driven uncertainty. Agents are evaluated primarily by accumulated earnings.


\begin{table*}[t]
\centering
\small
\setlength{\tabcolsep}{3pt}
\begin{tabular}{ccccccccccc}
\toprule
\multirow{2}{*}{Domain} & \multirow{2}{*}{Task} 
& \multicolumn{4}{c}{Single-agent} 
& \multicolumn{5}{c}{Multi-agent} \\
\cmidrule(lr){3-6} \cmidrule(lr){7-11}
& 
& CoT & Self-refine & Reflexion & AMEM 
& Debate & Discussion & DC & GEPA & ACE \\
\midrule

\multirow{1}{*}{Information Extraction}
& All $\uparrow$ 
&0.824 	&0.826 	&0.791 	& \best{0.857} 	&0.798 	&0.830 	&0.801 	&0.846 	& 0.853 \\
\midrule

\multirow{3}{*}{Numerical Calculation}
& Easy $\uparrow$ 
&0.840 	&0.825 	&0.816 	& \best{0.847} 	&0.776 	&0.624 	&0.834 	&0.836 	&0.825 \\
& Middle $\uparrow$ 
&0.730 	&0.697 	&0.712 	&0.711 	&0.588 	&0.527 	&0.700 	& \best{0.736} 	&0.712 \\
& Hard $\uparrow$ 
&0.500 	&0.500 	&0.500 	& \best{0.563} 	&0.375 	&0.375 	&0.438 	&\best{0.563} 	&0.500 \\
\midrule

\multirow{3}{*}{Domain Knowledge}
& Easy $\uparrow$ 
&0.925 	&0.906 	& \best{0.981} 	&0.925 	&0.906 	&0.925 	&0.793 	&0.925 	&0.925 \\
& Middle $\uparrow$ 
&0.817 	&0.819 	&0.808 	& 0.827 	&0.790 	& \best{0.829} 	&0.708 	&0.797 	&0.826 \\
& Hard $\uparrow$ 
&0.339 	&0.329 	&0.386 	&0.349 	&0.367 	&0.349 	&0.357 	&0.355 	& \best{0.464} \\
\midrule

\multirow{1}{*}{Complex Reasoning}
& All $\uparrow$ 
&0.576 	& \best{0.727} 	&0.515 	&0.636 	&0.606 	&0.636 	&0.606 	&0.697 	&0.697 \\
\midrule

\multirow{3}{*}{\parbox{3cm}{\centering Interactive\\Decision-making}}
& Consulting $\uparrow$ 
& 7.12 & 6.38 & 6.75 & 7.37
& 6.45 & 6.80 & 8.00 & \best{8.28} & 7.63 \\
& BeerGame $\downarrow$
& 2.54 & 4.67 & 5.10 & 2.25 & 2.96 & \best{1.94} & 5.61 & 2.57 & 3.09 \\
& EDT $\uparrow$ 
& 5.82 & 4.32 & 6.20 & \best{6.76} & 5.12 & 5.36 & 5.63 & 6.04 & 5.63\\
\bottomrule
\end{tabular}
\caption{Performance comparison of single-agent and multi-agent methods across different domains and tasks using DeepSeek-V3 as the backbone model. BeerGame results are reported as total costs in units of $\times 10^4$ (lower is better), while EDT results are reported in units of $10^6$ (higher is better).}

\label{tab:main_results}
\end{table*}
\section{Experiment Setup}

\subsection{Evaluated Agent Methods}
We evaluate a diverse set of representative LLM agent methods under the same task interface. The methods cover single-agent reasoning, iterative self-improvement, multi-agent collaboration, retrieval-augmented adaptation, reflective prompt optimization, and memory-based reasoning.

\paragraph{Single-agent reasoning and refinement.}
CoT~\cite{wei2022chain} serves as a basic reasoning baseline that elicits step-by-step solutions through chain-of-thought prompting. Self-Refine~\cite{madaan2023self} extends this paradigm by generating an initial answer, producing self-feedback, and refining the response. Reflexion~\cite{shinn2023reflexion} further incorporates memory of previous failures to guide subsequent reasoning.

\paragraph{Multi-agent collaboration.} 
Debate~\cite{du2023improving} uses adversarial multi-agent argumentation, while Discussion~\cite{li2023camel} uses cooperative multi-agent interaction to exchange reasoning traces and reach a shared answer. We implement both with fixed interaction rounds to preserve their core mechanisms while keeping evaluation cost comparable.

\paragraph{Adaptive, retrieval, and memory-augmented methods.}
DC (Dynamic Cheatsheet)~\cite{suzgun2026dynamic} retrieves task-relevant reasoning patterns from a curated knowledge base. GEPA~\cite{agrawal2026gepa} optimizes prompts through natural-language reflection over rollout trajectories and Genetic-Pareto search, enabling sample-efficient adaptation without updating model weights. ACE~\cite{zhang2026agentic} adaptively evolves a procedural playbook from previous successful trajectories. AMEM~\cite{xu2026mem} augments agents with external memory to store and retrieve context-relevant reasoning traces.

\paragraph{Resource and adaptation budgets.}
We preserve each method's native inference and adaptation configuration rather than enforcing an identical budget, which would alter methods that inherently require multiple calls, online memory, or prompt evolution. All methods use only information generated within the same benchmark run. Detailed call, token, history-access, and prompt-evolution statistics are provided in Appendix~\ref{app:resource_budgets}. 

\subsection{Evaluation Details}
We evaluate nine agent methods with DeepSeek-V3 and GPT-4.1, and seven representative methods with DeepSeek-V4-Pro and GLM-5.2. The main text reports results with DeepSeek-V3 and GPT-4.1, while additional results with DeepSeek-V4-Pro and GLM-5.2 are provided in Appendix~\ref{app:additional_backbones}. Within each backbone, all methods share the same task interface and task-specific protocol.
Foundational tasks use fixed-input answer generation and dataset-level metrics, including exact match, normalized string match, multiple-choice accuracy, numerical correctness, and code-execution-based verification. Consulting is evaluated through multi-turn interaction and rubric-based scoring over the full transcript and final recommendation. Beer Game uses 25-step sequential replenishment decisions and accumulated cost, where lower is better. EDT uses a scenario-level configuration followed by a rollout and accumulated earnings, where higher is better. For Consulting, the evaluated agent backbone and the LLM judge are kept from different model families. When GPT-4.1 is used as the evaluated backbone, DeepSeek-V3 is used as the judge; for the other evaluated backbones, GPT-4.1 is used as the judge. Additional task descriptions, evaluation metrics, simulation configurations, consulting prompts, and dataset statistics are provided in Appendix~\ref{sec:task_details}, Appendix~\ref{sec:consulting_prompts}, Appendix~\ref{app:beergame}, Appendix~\ref{app:edt_details}, and Appendix~\ref{sec:dataset_stats}.

\section{Experiment Results}
\begin{table*}[t]
\centering
\small
\setlength{\tabcolsep}{3pt}
\begin{tabular}{ccccccccccc}
\toprule
\multirow{2}{*}{Domain} & \multirow{2}{*}{Task} 
& \multicolumn{4}{c}{Single-agent} 
& \multicolumn{5}{c}{Multi-agent} \\
\cmidrule(lr){3-6} \cmidrule(lr){7-11}
& 
& CoT & Self-refine & Reflexion & AMEM 
& Debate & Discussion & DC & GEPA & ACE \\
\midrule

\multirow{1}{*}{Information Extraction}
& All $\uparrow$ 
& 0.863 & 0.852 & 0.837 & \best{0.871}
& 0.861 & 0.870 & 0.822 & 0.863 & 0.858 \\
\midrule

\multirow{3}{*}{Numerical Calculation}
& Easy $\uparrow$ 
& 0.860 & 0.851 & 0.850 & 0.931
& 0.851 & 0.864 & 0.922 & 0.943 & \best{0.963} \\
& Middle $\uparrow$ 
& 0.742 & 0.691 & 0.736 & 0.727
& 0.718 & 0.731 & \best{0.863} & 0.842 & 0.851 \\
& Hard $\uparrow$ 
& 0.500 & 0.500 & 0.500 & 0.500
& 0.500 & 0.500 & 0.500 & \best{0.563}  & 0.437 \\
\midrule

\multirow{3}{*}{Domain Knowledge}
& Easy $\uparrow$ 
& 0.887 & 0.887 & 0.944 & \best{0.962}
& 0.906 & 0.906 & 0.813 & 0.921 & 0.925 \\
& Middle $\uparrow$ 
& 0.866 & \best{0.877} & 0.870 & 0.861
& 0.866 & 0.873 & 0.814 & 0.825 & 0.829 \\
& Hard $\uparrow$ 
& 0.423 & 0.443 & 0.462 & 0.455
& 0.458 & \best{0.472} &  0.462 &  0.455 & 0.468 \\
\midrule

\multirow{1}{*}{Complex Reasoning}
& All $\uparrow$ 
& \best{0.758} & \best{0.758} & 0.727 & 0.727
& \best{0.758} & 0.727 & 0.713 & 0.727 & 0.697 \\
\midrule

\multirow{3}{*}{\parbox{3cm}{\centering Interactive\\Decision-making}}
& Consulting $\uparrow$ 
& 7.15 & 6.47 & 6.61 & 7.66
& 6.75 & 6.68 & \best{8.08} & 7.89 & 7.49 \\
& BeerGame $\downarrow$
& 5.36 & 4.73 & 4.62 & 3.82
& 3.17 & 5.47 & 3.38 & \best{3.09} & 4.15 \\
& EDT $\uparrow$ 
& 7.77 & 5.42 & 6.97 & 7.21
& 5.83 & 5.97 & 4.50 & \best{8.16} & 2.90 \\
\bottomrule
\end{tabular}
\caption{Performance comparison of single-agent and multi-agent methods across different domains and tasks using GPT-4.1 as the backbone model. BeerGame results are reported as total costs in units of $\times 10^4$ (lower is better), while EDT results are reported in units of $10^6$ (higher is better).}
\label{tab:gpt41_results}
\end{table*}

\subsection{Main Result}

Tables~\ref{tab:main_results} and~\ref{tab:gpt41_results} report results with DeepSeek-V3 and GPT-4.1, respectively. We organize the analysis around the foundational and interactive layers of EnterpriseBench.

\paragraph{Foundational tasks show clear difficulty effects.} 
The foundational layer exposes uneven capability limits across enterprise reasoning categories. Information Extraction shows relatively small variation across methods, indicating that evidence retrieval is less discriminative than computation-heavy or knowledge-intensive tasks. In contrast, Numerical Calculation drops sharply from easy to hard examples under both backbones, and Domain Knowledge follows a similar pattern. This shows that the reorganized QA layer does more than aggregate static datasets: it separates relatively stable document-level reasoning from more fragile numerical and specialized-knowledge reasoning.

\paragraph{Interactive tasks reveal different evaluation signals from static QA.}
Methods that perform well on static QA are not always the best on Consulting, Beer Game, or EDT. Under DeepSeek-V3, the leading method differs across the three interactive tasks: GEPA performs best on Consulting, Discussion achieves the lowest Beer Game cost, and AMEM obtains the highest EDT earnings. Under GPT-4.1, DC performs best on Consulting, while GEPA achieves the best Beer Game and EDT results. The relative ordering of the remaining methods also varies substantially across tasks and backbones. These results show that interactive enterprise decision-making cannot be fully assessed by fixed-input QA, because hidden-information dialogue, delayed-feedback control, and long-horizon planning require different agent capabilities. The ranking changes across backbones further indicate that agent effectiveness depends on the interaction between the agent design and the underlying model.

\subsection{Analysis of Interactive Tasks}

The three interactive tasks target different forms of enterprise decision-making: business diagnosis and recommendation under incomplete information in Consulting, supply-chain inventory control with delayed feedback in the Beer Game, and long-horizon enterprise decision-making in EDT. We analyze them separately to examine how agent behavior differs across these settings.
\paragraph{Consulting.}
Table~\ref{tab:consulting_dimensions} presents dimension-level results on the Consulting evaluation. Under DeepSeek-V3, GEPA achieves the highest overall score of 8.28 and leads in quantitative reasoning, business sense, and communication, while DC obtains the highest structure score and ranks second overall with 8.00. Under GPT-4.1, DC becomes the strongest method overall, achieving 8.08 and leading in structure, quantitative reasoning, and business sense. GEPA ranks second overall with 7.89 and achieves the highest communication score of 8.31. These results indicate that DC and GEPA are consistently strong, although their relative advantages depend on both the backbone and the evaluation dimension. The strongest methods perform well across multiple capabilities, indicating that Consulting requires agents to ask useful clarification questions, organize incomplete information, and present coherent recommendations. Self-Refine and Reflexion remain below CoT in overall score under both backbones, suggesting that iterative self-correction alone does not necessarily improve multi-turn business case solving. 
\paragraph{Beer Game.}
The Beer Game shows a different ranking. Since the metric is accumulated cost, lower values indicate better performance. Under DeepSeek-V3, Discussion achieves the lowest cost, reaching 1.94$\times 10^4$, followed by AMEM and CoT. In contrast, under GPT-4.1, GEPA achieves the lowest cost, while Discussion performs substantially worse. The reversal between DeepSeek-V3 and GPT-4.1 suggests that Beer Game performance depends on the fit between the agent design and the backbone model. This indicates that additional interaction is not always beneficial for delayed-feedback control.

\paragraph{EDT.}
EDT shifts the evaluation from step-by-step control to scenario-level planning. Agents choose workforce size, risk level, and project schedules before the simulator executes a full episode. Under DeepSeek-V3, AMEM achieves the highest accumulated earnings, reaching 6.76M, followed by Reflexion and GEPA. Under GPT-4.1, however, GEPA obtains the best EDT result, reaching 8.16M. The change in the leading method indicates that EDT performance is also backbone-dependent, with AMEM working best under DeepSeek-V3 and GEPA under GPT-4.1.

\begin{table*}[t]
\centering
\small
\setlength{\tabcolsep}{4.5pt}
\begin{tabular}{llccccccccc}
\toprule
\multirow{2}{*}{Backbone}
& \multirow{2}{*}{Dimension}
& \multicolumn{4}{c}{Single-agent}
& \multicolumn{5}{c}{Multi-agent} \\
\cmidrule(lr){3-6} \cmidrule(lr){7-11}
& & CoT & Self-refine & Reflexion & AMEM
& Debate & Discussion & DC & GEPA & ACE \\
\midrule

\multirow{5}{*}{DeepSeek-V3}
& Structure
& 7.26 & 6.54 & 6.90 & 7.40 & 6.60 & 6.99 & \best{8.15} & 8.08 & 7.68 \\
& Quant.
& 6.58 & 5.58 & 5.92 & 6.99 & 5.67 & 6.06 & 7.90 & \best{8.08} & 7.30 \\
& Business
& 7.30 & 6.56 & 6.93 & 7.44 & 6.73 & 7.01 & 8.15 & \best{8.40} & 7.70 \\
& Comm.
& 7.60 & 6.97 & 7.33 & 7.92 & 6.94 & 7.27 & 8.00 & \best{8.59} & 8.07 \\
& Overall
& 7.12 & 6.38 & 6.75 & 7.37 & 6.45 & 6.80 & 8.00 & \best{8.28} & 7.63 \\

\midrule

\multirow{5}{*}{GPT-4.1}
& Structure
& 7.32 & 6.70 & 6.81 & 7.71 & 6.89 & 6.85 & \best{8.22} & 7.71 & 7.53 \\
& Quant.
& 6.59 & 5.67 & 5.81 & 7.25 & 6.01 & 6.02 & \best{8.01} & 7.51 & 7.23 \\
& Business
& 7.33 & 6.64 & 6.78 & 7.75 & 7.02 & 6.86 & \best{8.28} & 8.04 & 7.53 \\
& Comm.
& 7.67 & 7.08 & 7.22 & 8.23 & 7.24 & 7.14 & 7.98 & \best{8.31} & 7.92 \\
& Overall
& 7.15 & 6.47 & 6.61 & 7.66 & 6.75 & 6.68 & \best{8.08} & 7.89 & 7.49 \\

\bottomrule
\end{tabular}
\caption{Performance of different agent methods on the Consulting task across multiple capability dimensions under DeepSeek-V3 and GPT-4.1 backbones.}
\label{tab:consulting_dimensions}
\end{table*}


\subsection{Validity and Reliability Analysis}
\label{sec:validity}

EnterpriseBench uses LLM-assisted capability categorization to reorganize QA tasks and LLM-based evaluation for open-ended consulting dialogues. We therefore conduct human audits and prompt robustness checks to assess the reliability of these components.

\begin{figure}[t]
    \centering
    \includegraphics[width=\linewidth]{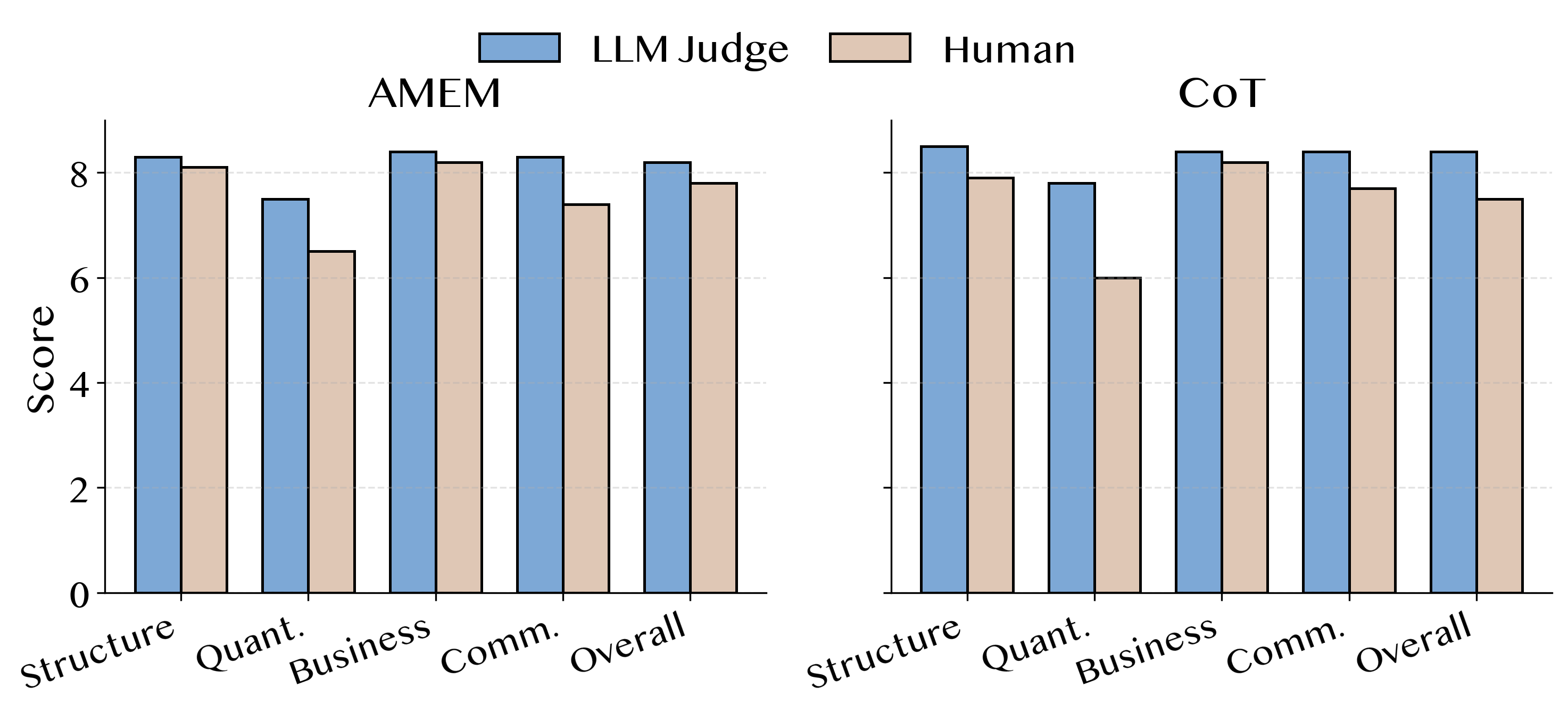}
\caption{
Comparison of human and LLM scores on 20 Consulting cases, averaging AMEM and CoT outputs per case.
    }
    \label{fig:consulting_human_audit}
\end{figure}

\paragraph{Human alignment of Consulting evaluation.}
We conduct a blind human audit on 20 consulting cases, evaluating both AMEM and CoT outputs for each case. Human evaluators are given the same information as the LLM judge: the original case text, the full dialogue transcript, and the four-dimension rubric. Since this audit was designed as an aggregate sanity check rather than a fine-grained agreement study, we compare average human and LLM scores at the method--dimension level.
As shown in Fig.~\ref{fig:consulting_human_audit}, the average human and LLM scores exhibit similar relative profiles across agent methods and evaluation dimensions, with a significant overall score correlation of $r=0.71$ ($p<0.001$) and an average gap of 8.5\%  relative to human scores. Human evaluators assign lower absolute scores, especially on quantitative reasoning and communication. This suggests that the LLM-based rubric is directionally aligned with human assessment, while quantitative reasoning may require stricter calibration. We further conduct an expanded transcript-level validation on 100 Consulting cases across five representative agent methods, yielding 500 method--case transcripts. Each transcript is independently evaluated by three human annotators. The expanded validation achieves an overall Pearson correlation of $r=0.887$, an 85.0\% within-one-point agreement rate, and a human ICC$(2,3)$ of 0.892. Full experimental details and dimension-level results are provided in Appendix~\ref{app:expanded_judge_validation}.

\begin{figure*}[t]
    \centering
    \includegraphics[width=1\linewidth]{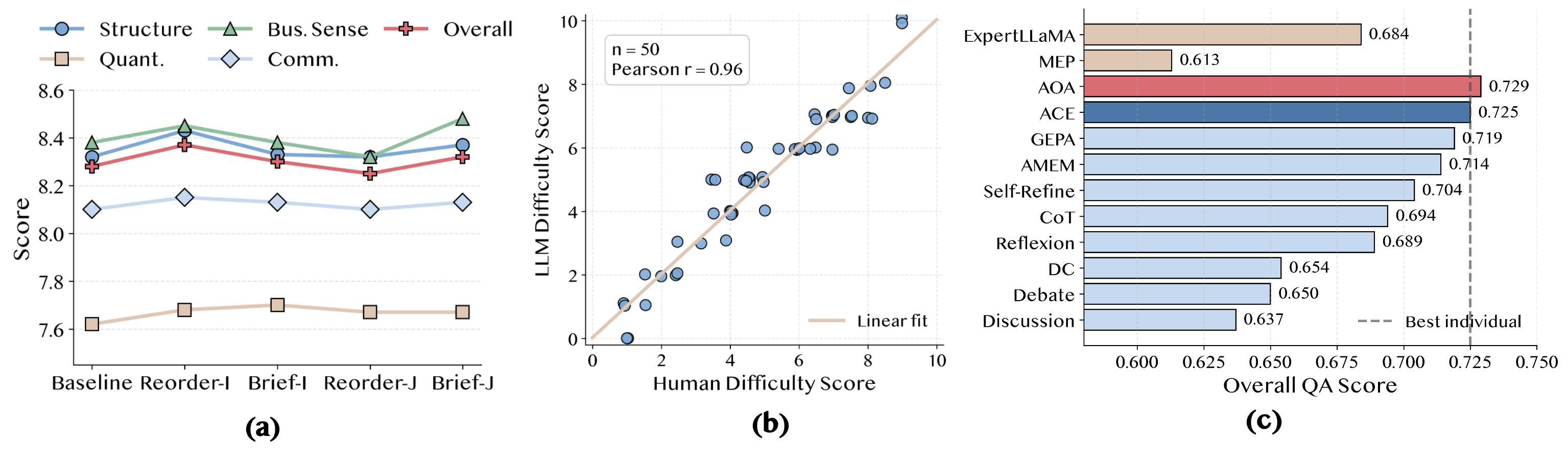}
    \caption{
    (a) Prompt robustness analysis for Consulting using ACE. Reorder-I and Brief-I denote interviewer prompt variants, while Reorder-J and Brief-J denote judge prompt variants. Scores remain stable across prompt variants, with the overall score changing by at most 0.12 points.
    (b) Human audit of difficulty annotation. Each point represents one audited task. Difficulty scores assigned by the LLM are strongly aligned with human ratings on the 0--10 scale, with Pearson $r=0.96$.
    (c) Performance comparison of AOA method against baseline approaches on QA tasks.
    }
    \label{fig:difficulty_human_llm_scatter}
\end{figure*}

\paragraph{Prompt robustness of Consulting evaluation.}
We also test whether the Consulting evaluation is sensitive to prompt wording. We construct variants of the interviewer and judge prompts by either reordering key instructions or simplifying their wording while preserving the same task requirements and scoring criteria. Here, I and J denote variants of the interviewer and judge prompts, respectively. As shown in Fig.~\ref{fig:difficulty_human_llm_scatter} (a), scores remain stable across all prompt variants: the maximum overall score deviation is 0.12 points, corresponding to a 1.45\% relative change from the baseline. These small deviations suggest that the Consulting results are not artifacts of a single prompt phrasing.

\paragraph{Reliability of difficulty annotation.}
We audit the difficulty annotation on a stratified sample of 50 tasks. Two domain experts independently score each task using the same difficulty anchors, and we use their averaged score as the human difficulty rating. As shown in Fig.~\ref{fig:difficulty_human_llm_scatter} (b), LLM-assigned difficulty scores are strongly aligned with human ratings, with Pearson $r=0.96$ and $p<0.001$. The mean absolute difference is 0.47 points on the 0--10 scale, and 94\% of samples differ by at most one point from the human score. These results indicate that the LLM-based difficulty annotations are closely aligned with expert human judgments.



\subsection{Adaptive Routing: A Proof of Concept}

The results above show that agent performance varies across task types and sample characteristics, motivating a question: can an agent system improve overall performance by assigning each instance to a suitable method? We study this question on the QA portion of EnterpriseBench, where samples have capability and difficulty annotations.
We first compare with two expert-style baselines, ExpertLLaMA~\cite{xu2023expertprompting} and Multi-expert Prompting~\cite{do2024multi}, which use expert roles or multiple expert perspectives. They achieve overall QA scores of 0.684 and 0.613, respectively, below the strongest general agent baselines. We then introduce AOA as a proof-of-concept adaptive agent selection method. AOA derives routing rules from execution traces and maps sample characteristics, such as capability category, difficulty level, table or code presence, and task source, to a selected base method. Rule synthesis is performed without using labels from held-out test instances, and details are provided in Appendix~\ref{sec:aoa_prompts}.
As shown in Fig.~\ref{fig:difficulty_human_llm_scatter}(c), AOA achieves an overall QA score of 0.729, slightly higher than ACE, the strongest individual method at 0.725. Although the gain is modest, this result suggests that adaptive agent selection can provide additional benefits over a fixed agent choice. Together with the expert-style baselines, this proof of concept indicates that expert-agent systems may benefit from selection mechanisms that account for task-specific conditions.

\section{Related Work}

Financial and business-domain benchmarks mainly evaluate static report understanding, numerical reasoning, domain knowledge, financial QA, and information structuring \cite{chen2021finqa, reddy2024docfinqa, krumdick2024bizbench, xie2024finben, zhang2025xfinbench}. General agent benchmarks study tool use, web interaction, embodied environments, and simulations \cite{qin2024toolllm, zhou2024webarena, deng2023mind2web, shridhar2020alfworld, wang2022scienceworld}, but are not designed around enterprise strategic reasoning. EnterpriseBench complements these efforts with enterprise-focused QA and interactive tasks for information-seeking dialogue, delayed-feedback control, and long-horizon planning. See Appendix~\ref{sec:extended_related_work}.

\section{Conclusion}
We present \textbf{EnterpriseBench}, a benchmark for evaluating LLM agents from foundational enterprise reasoning to interactive decision-making. It combines reorganized enterprise and financial QA datasets with three interactive tasks: Consulting, Beer Game, and Enterprise Digital Twin. Experiments with nine agent methods and four backbone models show that static QA performance does not reliably predict interactive decision-making, and that current agents still face distinct challenges across static, interactive, and simulation-based enterprise tasks.

\section*{Limitations}
EnterpriseBench has several limitations. First, its interactive tasks are implemented through controlled benchmark environments. This design enables reproducible evaluation of information seeking, delayed feedback, and long-horizon planning, but it does not fully capture all social, organizational, and institutional factors in real enterprise decision-making. Second, the current benchmark focuses on text-based enterprise scenarios and does not yet cover multimodal evidence such as slides, spreadsheets, emails, dashboards, or meeting transcripts, which are common in enterprise workflows. Third, the Consulting, Beer Game, and EDT tasks cover representative forms of enterprise decision-making, but they do not exhaust all industries, organizational scales, or decision types. Extending EnterpriseBench to more specialized sectors and richer operational settings is an important direction for future work. Finally, our experiments cover nine representative agent methods and four backbone models; future studies may include additional agent architectures, proprietary enterprise systems, and human-AI collaboration settings. Because EnterpriseBench evaluates agents in enterprise decision-making scenarios, its results should not be interpreted as evidence that LLM agents are ready for autonomous deployment in high-stakes business, financial, or operational decisions.

\section*{Acknowledgments}
This work was supported by the National Key Research and Development Program of China (Grant No. 2024YFC3307603). This work was also supported by the Zhongguancun Academy (Grant No. C20250201).

\bibliography{custom}

\appendix

\section{Appendix}

\subsection{Detailed Task Descriptions and Metrics}
\label{sec:task_details}

\paragraph{Structured Reasoning (QA)}
We evaluate our agents on a comprehensive set of financial and logical reasoning tasks covering four key domains: Information Extraction, Numerical Calculation, Domain Knowledge, and Complex Reasoning.
\textbf{Metrics:} We employ task-specific evaluation protocols to ensure rigorous assessment:
\begin{itemize}
    \item \textbf{Program Synthesis:} For tasks requiring code generation, we use a sandboxed Python environment. Correctness is determined by a hybrid tolerance model: using an absolute tolerance of $10^{-6}$ for values near zero and a relative tolerance of 0.01 for larger magnitudes.
    \item \textbf{Strict Quantity Extraction:} For tasks like TAT-QA, we enforce a zero-relative-tolerance policy. Predictions must match ground truth within a strict absolute epsilon of $10^{-6}$, or match the normalized text span exactly.
    \item \textbf{Numerical Formula Evaluation:} For the \textit{formula} task, agents must perform multi-step financial calculations. Correctness is verified by extracting the final numerical answer (supporting JSON-formatted or free-text responses) and performing exact or standardized floating-point comparison.
    \item \textbf{Multiple Choice \& Tagging:} For knowledge tasks, we use regex to extract option letters. For financial tagging (\textit{finer}), we evaluate the exact match of comma-separated entity labels, supporting both raw string and evaluated numerical comparisons.
\end{itemize}

\paragraph{Consulting Case Analysis}
This task simulates management consulting interviews where agents must solve complex business problems.
\textbf{Adaptation:} Each agent method is instantiated as a ``candidate'' who receives a detailed business case and must provide structured recommendations.
\textbf{Metrics:} Evaluation is performed across four dimensions: \textit{Structure}, \textit{Quantitative Reasoning}, \textit{Business Sense}, and \textit{Communication}, with an \textit{Overall} score computed as the average.

\paragraph{Beer Game (Serious Game)}
A multi-turn supply chain simulation where agents manage inventory levels across multiple tiers.
\textbf{Adaptation:} Agents take on roles (e.g., retailer) and must make replenishment decisions based on dynamic market demand and lead times.
\textbf{Metrics:} The performance is measured by \textit{Total Cost}, which includes inventory holding costs and backlog penalties. A lower cost indicates superior strategic planning.

\paragraph{Enterprise Digital Twin (EDT)}
A high-fidelity simulation of an entire enterprise ecosystem requiring long-term strategic decision-making.
\textbf{Adaptation:} Each agent jointly decides project selections and workforce allocations, which are evaluated through BPTK (Business Process Tool Kit)  simulation environment. 
\textbf{Metrics:} The primary metric is \textit{Accumulated Earnings}, representing the total profit generated by the enterprise during the simulation period.

\subsection{Agent-based Classification Prompts}
\label{sec:reclass_prompts}

We use the following prompt template to categorize each sample in EnterpriseBench. The variables \texttt{\{task\_name\}} and \texttt{\{question\_block\}} are dynamically populated during the classification process.

\begin{lstlisting}
You are an expert in question quality review and capability evaluation.

Do NOT solve the question. Your task is to analyze what the question is testing.

TaskName: {task_name}

### Capability Categories
Select ONLY ONE primary capability from the following four categories:

1. Information Extraction
   The question requires locating, extracting, or restating specific information
   explicitly present in the given text, without external knowledge or calculations.

2. Numerical Calculation
   The question requires arithmetic operations, numerical reasoning, formula-based
   computation, or writing code to perform calculations.

3. Domain Knowledge
   The question primarily tests specialized knowledge itself, such as:
   - Finance, accounting, XBRL, or US GAAP concepts and tags
   - Financial formulas or accounting standards
   - CFA exam questions focused on knowledge recall or understanding
   - MMLU questions in business ethics, microeconomics, or professional accounting
     that assess knowledge rather than decision-making

4. Complex Reasoning
   The question requires judgment or decision-making under constraints, such as:
   - Scenario-based CFA exam questions
   - MMLU questions involving business ethics, microeconomics, or professional
     accounting that ask what should be done in a given situation

### Decision Rules (IMPORTANT)
Use these rules to avoid confusion between Information Extraction and Numerical Calculation:

- Information Extraction ONLY IF:
  - The final answer can be directly copied from the provided text/table/verbatim span, AND
  - NO arithmetic/computation is required.

- Numerical Calculation IF ANY computation is needed, even if very simple, including but not limited to:
  - subtraction / difference / net change (e.g., "2015 - 2014")
  - addition / sum / total across years or rows
  - division / ratio / percentage / percent change / growth rate
  - max/min/average over a set of numbers
  - any formula-based computation, or any need to write/execute code

Examples:
- If the context contains two values and the question asks "net change" / "difference" / "by what percentage",
  this is Numerical Calculation (even though the values are extracted from the context).
- If the question asks "What is X in 2017?" and X is explicitly stated as a single value in the table,
  this is Information Extraction.

### Your Task
Analyze the following question and provide:
1. The primary capability category (choose exactly one)
2. A difficulty score from 0 to 10 (0 = trivial, 10 = extremely hard)
3. A brief justification explaining both the capability classification and
   the difficulty level

### Output Format (strictly follow):
Capability: <Information Extraction | Numerical Calculation | Domain Knowledge | Complex Reasoning>
DifficultyScore: <0-10>
Reasoning: <Up to 4 sentences>

### Scoring Guidance (IMPORTANT)
Be critical and use the full 0-10 range as much as possible. Avoid clustering scores in 0-5.
Assign higher scores (6-10) to genuinely challenging questions (long/complex context, multi-step computation, tricky domain knowledge, ambiguity, or decision-making).
Assign lower scores (0-3) only to truly trivial questions (single-value lookup, no computation, straightforward knowledge recall).

### Special Difficulty Rules for Label-Selection / Classification Tasks (IMPORTANT)
Some tasks look short but are hard because the model must choose the correct label from a large, confusing label space
(e.g., US-GAAP XBRL tag selection, fine-grained schema mapping).

- If the input contains a long list of candidate labels/tags (dozens to hundreds), difficulty should be HIGH (often 7-10),
  even if the sentence is short, because many labels are semantically overlapping.
- If there are multiple independent sub-questions in one sample (e.g., "answer the following 4 questions"), difficulty increases.
- If the correct label is not explicitly named and requires semantic interpretation to disambiguate between similar labels, difficulty increases.
- Reserve 0-3 only for cases where the label set is tiny and the mapping is obvious/unambiguous.

### Question to Analyze:
{question_block}
\end{lstlisting}

The classification is performed using the \texttt{deepseek-v3} model to ensure consistency across the entire benchmark.

\subsection{AOA Experience Extraction Prompts}
\label{sec:aoa_prompts}

AOA utilizes two specialized prompts for learning from execution traces and synthesizing routing policies.

\paragraph{Trace-based Experience Extraction Prompt}
This prompt is used to analyze individual agent reasoning traces and extract actionable experience bullets.

\begin{lstlisting}
# AOA Trace-based Experience Extractor (Per-sample)
You are an expert evaluator of LLM-agent reasoning traces.

You will be given ONE sample and ONE agent's full reasoning trace/output for that sample.
Your task is to extract ONE reusable experience item that can help:
- improve future usage of this agent, and/or
- decide when to route similar samples to a different agent.

## Hard constraints
- Output **JSON only**. No markdown. No extra text.
- Be concrete: reference the failure/success mode, not generic advice.
- If the sample is incorrect, diagnose the likely cause (format mismatch, sign error, unit conversion, missing table lookup, etc.).
- If the sample is correct, extract what made it work (e.g., robust checks, careful parsing, etc.).

## Output JSON schema (strict)
{
  "bullet": "<one actionable experience sentence, English>",
  "tags": {
    "agent_method": "<string>",
    "task_name": "<string>",
    "capability": "<string>",
    "difficulty_bucket": "<easy|middle|hard|NA>"
  },
  "outcome": "<correct|incorrect>",
  "diagnosis": "<short root-cause / success-factor>",
  "routing_hint": {
    "prefer_agent": "<agent_name or empty>",
    "avoid_agent": "<agent_name or empty>",
    "when": "<short condition description>"
  },
  "confidence": "<high|medium|low>"
}
\end{lstlisting}

\paragraph{Meta-level Experience Synthesizer Prompt}
This prompt is used to aggregate individual experiences into a unified, executable routing policy.

\begin{lstlisting}
# AOA Meta Experience Synthesizer
You are an expert in LLM-agent routing and evaluation.

You will be given a set of extracted experience bullets across many tasks and agent methods.
Your task is to produce a **meta-level routing playbook** that helps an AOA router decide
which agent to use for a new sample.

## Hard constraints
- Output **JSON only** (no markdown, no extra text).
- Your routing_policy rules must be executable based on: task_name, capability, difficulty_bucket, and simple text features.
- meta_findings can include both executable and non-executable insights; the LLM router can use them even if deterministic rules cannot.
- Prefer simple, robust rules. Avoid overfitting to tiny evidence.
- Use a conservative default_agent.

## Output JSON schema (strict)
{
  "meta_findings": [
    {
      "id": "M1",
      "summary": "<one sentence>",
      "evidence": "<short evidence based on provided bullets/stats>",
      "confidence": "<high|medium|low>"
    }
  ],
  "routing_policy": {
    "default_agent": "<agent_name>",
    "tie_breaker": "prefer_default|prefer_simpler|prefer_ace",
    "min_margin": 0.01,
    "rules": [
      {
        "when": {
          "task_name": "<name|ALL>",
          "capability": "<name|ALL>",
          "difficulty_bucket": "<easy|middle|hard|NA|ALL>",
          "feature_conditions": {
            "has_table": "<true|false|omit>",
            "has_code": "<true|false|omit>",
            "num_numbers_min": "<number|omit>",
            "num_numbers_max": "<number|omit>",
            "context_chars_min": "<number|omit>",
            "context_chars_max": "<number|omit>",
            "question_chars_min": "<number|omit>",
            "question_chars_max": "<number|omit>",
            "table_row_estimate_min": "<number|omit>",
            "table_row_estimate_max": "<number|omit>"
          }
        },
        "choose": "<agent_name>",
        "rationale": "<one sentence>",
        "confidence": "<high|medium|low>"
      }
    ]
  }
}
\end{lstlisting}

\subsection{Consulting Prompts} 
\label{sec:consulting_prompts}
\paragraph{Interviewer prompt} The interviewer prompt is designed to guide the LLM interviewer to conduct a realistic consulting-style case interview based on predefined consulting cases. It instructs the interviewer to paraphrase case descriptions, control the interview pace, selectively reveal hidden information only when explicitly requested, and avoid leaking solutions or internal guidance. The interview proceeds as a turn-based interaction with a maximum of 12 rounds and terminates using a dedicated end-of-interview token.

\begin{lstlisting}
You are the interviewer in a consulting-style case interview with an LLM candidate.

Each turn you receive:
1) These instructions.
2) The full text of ONE case (problem/background, any sections such as
   "Information to be provided if requested", "If asked for market information",
   "Hints", "Key questions", "Analysis", "Possible recommendations / approaches",
   "Solution", "Case wrap-up", "Interviewer notes", etc.).
3) The chat history so far.

Your job is ONLY to produce the next interviewer message.

1. What you may reveal
- The problem statement, scenario description, and general background are information
  the candidate is allowed to know.
- Sections like "Information to be provided if requested", "If asked for ...",
  "Further information", "Hints" or similar are GATED facts.
- Sections like "Key questions", "Possible recommendations / approaches", "Analysis",
  "Solution", "Case wrap-up", "Interviewer notes" are INTERNAL GUIDANCE ONLY.

Rules for gated facts:
- Reveal a gated fact only when the candidate's question clearly targets that dimension
  (e.g., market size, growth, costs, customers, competition, operations, risks).
- Reveal one logical piece at a time, not the whole section at once.
- Never quote or expose "Solution", "Analysis", "Case wrap-up",
  "Possible recommendations / approaches" or "Interviewer notes" directly.
  Use them only to decide what to probe and which facts matter.

Do NOT invent or assume new facts beyond the case. If the candidate asks for
information that is not in the case and not covered by any gated section, say that the
case does not provide that detail and invite them to proceed with reasonable assumptions
or move to another relevant angle.

2. How to run the interview

First turn:
- Briefly set up the situation in your own words (1-3 sentences).
- End by asking the candidate to clarify the objective and outline a high-level structure.

Later turns:
- Read the latest candidate answer and the history.
- Answer their concrete questions using only allowed and already-unlocked information
  (plus any newly unlocked gated facts).
- Ask ONE focused follow-up at a time, pushing them toward structured business
  reasoning (e.g., profitability, market / customer / competition, operations, risks).
- Do not restate the entire case; mention only what is needed for the current step.

Pacing and depth:
- Use the case length hint (e.g. "Short 15 Minutes", "Medium 30 Minutes",
  "Long 45 Minutes") and the conversation so far to manage depth:
  * Short cases: aim for at least 3-4 candidate answers before closing.
  * Medium cases: aim for 5-7 candidate answers
  * Long cases: aim for 7-10 candidate answers with deeper quantitative or conceptual work.
- If the candidate keeps asking for more data without analyzing, gently redirect them to:
  (a) summarize what they know, and (b) propose a structure or hypothesis BEFORE you give more data.
- If the candidate is stuck, you may give a small hint or suggest one missing dimension,
  but do NOT present a full framework or full solution.

3. Ending the case

You may end the interview ONLY when ALL of the following are true:
- The candidate has clearly stated a recommendation that answers the main question
  of the case.
- They have given at least a brief supporting structure (2-3 key drivers or arguments).
- You have given short, high-level feedback and, if appropriate, added one or two
  important missing points.

Ending protocol:
- NEVER end the interview in your very first message.
- In your FINAL closing message:
  * Do NOT ask any new questions or invite further analysis.
  * Optionally give concise feedback and highlight key drivers.
  * Append the exact token {INTERVIEW_END_TOKEN} as the VERY LAST characters.
- In all earlier messages you MUST NOT output {INTERVIEW_END_TOKEN}.

4. Style and constraints

- Speak as a professional human interviewer: concise, neutral, business-like.
- Ask at most one or a small cluster of closely related questions per turn.
- Never mention "case text", "sections", "gated information", "solutions", or any
  internal labels; to the candidate you are simply an interviewer.
- Use only information from the case text and the chat history; do not bring in
  outside knowledge.
- Keep each interviewer message concise: at most 500 words in each turn. Do not write long essays.

Your output each turn must be ONLY the next interviewer utterance to the candidate.
"""
\end{lstlisting}

\paragraph{Judge prompt}  The judge prompt is designed to provide a consistent and fine-grained evaluation of agent performance in the consulting task. Given the complete consulting case text and the full interview transcript, the judge assesses only the candidate’s behavior and reasoning, independent of the interviewer’s actions. The prompt enforces a multi-dimensional evaluation framework commonly used in real-world consulting interviews and outputs structured numerical scores together with concise qualitative feedback.

The evaluation dimensions are as follows:

\begin{itemize}
    \item \textbf{Structure}: Evaluates whether the candidate clearly understands the problem and proposes a coherent, logically organized, and adaptable problem-solving structure.

    \item \textbf{Quantitative Reasoning}: Assesses the candidate’s ability to request, interpret, and apply numerical information to derive meaningful quantitative insights.

    \item \textbf{Business Sense}: Measures whether the candidate identifies key business drivers and trade-offs and provides commercially reasonable conclusions and risk-aware recommendations.

    \item \textbf{Communication}: Examines the clarity, conciseness, and professionalism of the candidate’s communication throughout the interview.

    \item \textbf{Overall}: Provides a holistic judgment of the candidate’s suitability for a consulting role, beyond a simple aggregation of individual dimension scores
\end{itemize}

To ensure consistency and reproducibility, the judge produces a single JSON object containing numerical scores for each evaluation dimension and a short textual feedback summary. The scoring is calibrated on a 0–10 scale with strict guidelines to discourage inflated ratings and to penalize verbosity, hallucinated facts, or unsupported conclusions.

\begin{lstlisting}
You are a senior consulting interviewer evaluating the performance of a CANDIDATE
in a case interview.

You will receive:
- case_text: the full written case (problem, background, solution, etc.);
- transcript_text: the complete dialogue between INTERVIEWER and CANDIDATE,
  in chronological order. Each line clearly indicates who is speaking.

Your job is to assess ONLY the CANDIDATE, not the interviewer.

Evaluate the candidate along FOUR dimensions plus an overall score:

1) structure (0-10)
   - How well does the candidate understand and restate the problem and objective?
   - Do they propose a clear, logical, and MECE-enough structure or approach early on?
   - Do they use hypothesis-driven thinking and adjust their structure as new information appears?

2) quant (0-10)
   - Does the candidate ask for the right type of information or data when needed?
   - Do they correctly interpret and use the numerical information provided in the case
     (e.g., doing rough calculations, sanity checks, comparisons)?
   - Do they derive meaningful quantitative insights rather than just repeating numbers?

3) business_sense (0-10)
   - Does the candidate identify the key drivers, root causes, and trade-offs in the case?
   - Are their conclusions and recommendations commercially reasonable and consistent
     with the information given?
   - Do they recognize important risks/uncertainties and, when appropriate, suggest
     sensible next steps or mitigations?

4) communication (0-10)
   - Is the candidate's communication clear, concise, and well-structured?
   - Do they signpost their thinking (e.g., "first/second/third") without being verbose?
   - Do they interact professionally with the interviewer, responding to questions,
     picking up on hints, and keeping a natural case-interview flow?

In addition, provide:

5) overall (0-10)
   - Your holistic judgment of the candidate's performance on this case.
   - This is NOT just an arithmetic average; it reflects whether you would be
     comfortable recommending this candidate for a consulting role.

Scoring guidelines (be strict and well-calibrated across many cases):
- 0-2: very weak (almost no useful contribution or completely off-track).
- 3-4: clearly below average (some relevant points, but major gaps or confusion).
- 5-6: average candidate (generally reasonable but shallow, incomplete, or inconsistent).
- 7: above average (solid performance with notable but fixable weaknesses).
- 8: very strong (consultant-level performance with only minor issues).
- 9-10: truly exceptional (outstanding on almost all dimensions; reserve for rare cases).

Additional rules:
- If the candidate barely speaks, never proposes a clear structure, or never gives a
  concrete recommendation, most scores should be in the 0-3 range.
- Do NOT reward verbosity alone; reward clear, structured, business-relevant thinking.
- Penalize hallucinated facts that contradict or go beyond the case_text.

Output format:
Return ONLY a single valid JSON object with this exact schema:
{
  "structure": float,
  "quant": float,
  "business_sense": float,
  "communication": float,
  "overall": float,
  "feedback": string
}
No extra text before or after the JSON.

\end{lstlisting}
\subsection{Beer Game Configuration}
\label{app:beergame}

\paragraph{Configuration parameters} The Beer Game simulation follows a discrete time setting with a fixed horizon of 25 time steps. Customer demand is observed only by the retailer and follows a stepwise demand script: demand remains at a low level of 100 units during the initial phase and increases to a high level of 400 units starting from week 2. This sudden demand shift introduces non-stationarity and tests the agent’s ability to adapt to changing market conditions.

Information and material flows are subject to delays. Orders placed by downstream agents are transmitted upstream with a one-week information delay, while physical shipments experience a two-week delivery delay. The system is initialized in a steady state with a target inventory level of 400 units to avoid transient effects at the beginning of the simulation.

Inventory dynamics incur explicit economic costs. Each unit of inventory held generates a holding cost of 0.5 per time step, while each unit of unmet demand (backorder) incurs a higher penalty of 1.0, reflecting the greater economic impact of stockouts. In addition, a minimum inventory cost of 200 is imposed to model fixed operational expenses independent of inventory fluctuations.

In all experiments, the evaluated agent controls the retailer role, which is closest to customer demand and therefore most exposed to demand uncertainty. All other supply-chain roles (wholesaler, distributor, and factory) follow predefined equation-based policies.

\paragraph{Opponent policies}
Non-controlled agents in the supply chain follow fixed equation-based ordering rules. Under the \textit{typical} policy, the order quantity at time step $t$ is determined by an inventory and backlog correction rule:
\begin{equation}
    q_t = d_t + \alpha \left(I^{*} - I_t\right) + \beta B_t, \label{eq:typical}
\end{equation}

where $d_t$ denotes the observed demand, $I_t$ is the current inventory level, $I^{*}$ is the target inventory, and $B_t$ represents the backlog. The parameters $\alpha$ and $\beta$ control the adjustment speed toward the target inventory and backlog compensation, respectively. And in our experiment we set $\alpha = 1$ and $\beta = 1$ as fix parameters.

Under the \textit{smoothing\_4} policy, demand is first smoothed using a four-step moving average:
\begin{equation}
    \tilde{d}_t = \frac{1}{4} \sum_{k=0}^{3} d_{t-k}, \label{eq:smoothing_dt}
\end{equation}

and the smoothed demand $\tilde{d}_t$ is then substituted for $d_t$ in the ordering
rule:
\begin{equation}
    q_t = \tilde{d}_t + \alpha \left(I^{*} - I_t\right) + \beta B_t, \label{eq:smoothing_main}
\end{equation}

\subsection{Enterprise Digital Twin (EDT) Implementation Details}
\label{app:edt_details}

This appendix provides low-level execution details of the Enterprise Digital Twin (EDT) task, complementing the main text description. We focus on (i) scenario specification, (ii) step-wise simulation dynamics, (iii) project life-cycle and stochastic events (extension and follow-on), (iv) firm-level accounting and metrics, and (v) the evaluation protocol used in our benchmark. 

Here is a brief version of the scenario applied in our task, and we use it as an example for illustrating the mechanism of EDT environment. The details of this task scenario can be found in our codes.

\begin{lstlisting}
"scenarios": {
  "interactive": {
  
    "runspecs": {
      "starttime": 1,
      "stoptime": 96,
      "dt": 1
    },
    
    "properties": {
      "revenue_risk_level": {
        "type": "Double",
        "value": 0.5
      },
      "fixed_cost": {
        "type": "Double",
        "value": 20000.0
      }
    },
    
    "agents": [
      {
        "name": "consultant",
        "count": 1,
        "properties": {
          "name": {
            "type": "String",
            "value": "Consultant 1"
          },
          "salary": {
            "type": "Double",
            "value": 6000.0
          },
          "workplace_cost": {
            "type": "Double",
            "value": 2000.0
          }
        }
       }, 
       
       ... # 11 other consulatant agents

       {
        "name": "project",
        "count": 1,
        "properties": {
          "name": {
            "type": "String",
            "value": "Project 1: Core Upgrade"
          },
          "contracted_effort": {
            "type": "Double",
            "value": 70.0
          },
          "contracted_probability": {
            "type": "Double",
            "value": 1.0
          },
          "extension_probability": {
            "type": "Double",
            "value": 0.25
          },
          "extension_effort": {
            "type": "Double",
            "value": 10.0
          },
          "follow_on_probability": {
            "type": "Double",
            "value": 0.1
          },
          "is_follow_on": {
            "type": "Boolean",
            "value": false
          },
          "deadline": {
            "type": "Double",
            "value": 30.0
          },
          "consultants": {
            "type": "Double",
            "value": 2.0
          },
          "start_time": {
            "type": "Double",
            "value": 1.0
          },
          "billing_rate": {
            "type": "Double",
            "value": 16000.0
          }
        }
       },

       ... # 9 other project agents
       
    ]
\end{lstlisting}

\subsubsection{Scenario specification}
\label{app:edt_scenario_spec}

An EDT episode is defined by a JSON scenario under a scenario manager (e.g., smEDT), consisting of three top-level blocks: runspecs, properties, and agents. The runspecs block defines the discrete simulation horizon via starttime, stoptime, and dt. In our benchmark implementation, the simulator advances by discrete step calls until termination, and stoptime acts as a step limit . The scalar dt is used as a per-step scaling factor for both work delivery and cost accrual.

The properties block contains firm-level parameters, most importantly the global fixed\_cost and revenue\_risk\_level. The agents block instantiates a multiset of consultant agents and project agents, plus a controlling component that aggregates flows into evaluation metrics.

A tested agent does not act during the episode. Instead, it outputs a compact scenario-level decision schema
\(\{C, R, P\}\), where \(C\) is the retained number of consultants, \(R\) is the global revenue\_risk\_level, and \(P\) encodes per-project acceptance and start/deadline windows. This schema is applied as a constrained transformation to a template scenario to produce a materialized scenario JSON. Non-controllable template fields (e.g., salary, workplace cost, and baseline project parameters) remain unchanged. The simulator then executes the materialized scenario for the full horizon and returns step-wise and terminal outcomes.

\subsubsection{Consultant dynamics and capacity constraints}
\label{app:edt_consultant_dynamics}

Each consultant agent is characterized by two per-step cost parameters: salary and workplace\_cost. These costs are counted every step regardless of utilization and same for every consultant in our experiment settings. Let \(N_c\) denote the retained number of consultants (set by the tested agent), s denote the salary and w denote the workplace cost. At each step \(t\), the total consultant operating cost contribution is
\begin{equation}
\mathrm{Cost}^c_t \;=\; N_c \cdot (\mathrm{s} + \mathrm{w}) \cdot dt.
\end{equation}

Consultants provide the only labor capacity for project delivery. At any step, a consultant can contribute to at most one project. Moreover, each project \(j\) has an integer staffing requirement \(\mathrm{req}_j\) (scenario field consultants) that caps concurrent workers on that project: at any step, no more than \(\mathrm{req}_j\) consultants can deliver effort to project \(j\). The simulator enforces sticky assignment: once a consultant begins working on a project, it remains assigned to that project until the project finishes its current workload (base scope and any triggered extension), after which the consultant becomes available for reassignment. This stickiness induces non-trivial opportunity costs: starting a long project early can lock capacity and delay higher-margin projects.

Per-step work delivery is modeled in units of \emph{effort}. For a working consultant, delivered effort per step is scaled by \(dt\). Let \(k_{j,t}\) be the number of consultants actually working on project \(j\) at step \(t\), with
\(0 \le k_{j,t} \le \mathrm{req}_j\).
Then the delivered effort to project \(j\) at step \(t\) is upper bounded by
\begin{equation}
e_{j,t} \;\le\; k_{j,t}\cdot dt.
\end{equation}
This definition is consistent with the main-text statement that a consultant produces one unit of effort per step when \(dt=1\) in our settings.

\subsubsection{Project state}
\label{app:edt_project_dynamics}

Each project \(j\) is parameterized by:
\begin{description}
  \item[\texttt{contracted\_effort:}] base scope \(\mathrm{Effort}^{\mathrm{base}}_j\)
  \item[\texttt{billing\_rate:}] per-unit revenue \(\mathrm{Rate}_j\)
  \item[\texttt{consultants:}] staffing cap \(\mathrm{req}_j\)
  \item[\texttt{start\_time:}] start time \(\mathrm{start}_j\)
  \item[\texttt{deadline:}] deadline \(\mathrm{dead}_j\)
  \item[\texttt{contracted\_probability:}] reliability \(\pi^{\mathrm{base}}_j\)
  \item[\texttt{extension\_probability:}] the probability that a project will extend and requires more effort.
  \item[\texttt{extension\_effort}] the required effort in extension scope \(\mathrm{Effort}^{\mathrm{ext}}_j\)
  \item[\texttt{follow\_on\_probability}] follow-on probability of a project \(\pi^{\mathrm{fo}}_j\)
\end{description}

A project is active only within its permissible window. Work can accrue only when \(t \ge \mathrm{start}_j\) and \(t \le \mathrm{dead}_j\) and the episode has not terminated. Projects maintain a remaining workload state \(E_{j,t}\) (initialized to \(\mathrm{Effort}^{\mathrm{base}}_j\)). Given delivered effort \(e_{j,t}\), the update is
\begin{equation}
E_{j,t+1} \;=\; \max\bigl(0,\; E_{j,t} - e_{j,t}\bigr).
\end{equation}

Revenue is generated proportional to delivered effort. The instantaneous revenue contribution from project \(j\) at step \(t\) is
\begin{equation}
\mathrm{Rev}_{j,t} \;=\; e_{j,t} \cdot \mathrm{Rate}_j,
\end{equation}
and total step revenue is
\begin{equation}
\mathrm{Rev}_t \;=\; \sum_j \mathrm{Rev}_{j,t}.
\end{equation}
If the project reaches its deadline with \(E_{j,t} > 0\), the unfinished portion is not deliverable after \(\mathrm{dead}_j\), meaning it cannot produce further revenue within that project instance.

\subsubsection{Extensions and follow-on projects}
\label{app:edt_stochastic_events}

EDT introduces two stochastic mechanisms that can create additional revenue opportunities while consuming capacity and increasing uncertainty: extensions (scope growth) and follow-on projects.

\paragraph{Extensions.}
If a project completes its base scope strictly before its deadline (i.e., \(E_{j,t}=0\) at some \(t < \mathrm{dead}_j\)), an extension event may trigger. Let \(g(R)\) be the risk gating function induced by the global revenue\_risk\_level \(R\in[0,1]\). Operationally, the extension triggers when a project-level draw passes a threshold that depends on both \(\pi^{\mathrm{ext}}_j\) and \(R\). When triggered, the project remaining workload is increased by \(\mathrm{Effort}^{\mathrm{ext}}_j\):
\begin{equation}
E_{j,t} \leftarrow E_{j,t} + \mathrm{Effort}^{\mathrm{ext}}_j.
\end{equation}
The project then continues to consume consultant capacity and can generate additional revenue as the extension workload is delivered, subject again to the deadline and episode termination.

\paragraph{Follow-on projects.}
At a project's deadline step \(t=\mathrm{dead}_j\), a follow-on opportunity may be instantiated. As with extensions, instantiation is gated by the global risk level \(R\) and the project parameter \(\pi^{\mathrm{fo}}_j\). A follow-on project is created as a new project agent with is\_follow\_on = true. It inherits the primary economic structure of its parent project (e.g., similar staffing requirement and billing rate) while suppressing further follow-on chaining (follow-on probability set to zero), preventing infinite cascades. The follow-on has its own start time after creation and competes for the same consultant pool.

These stochastic mechanisms create non-linear portfolio effects. High \(R\) increases the chance of extensions and follow-ons, potentially improving earnings but also locking capacity and raising uncertainty; low \(R\) yields more predictable revenue trajectories but less upside.

\subsubsection{Firm-level accounting and returned metrics}
\label{app:edt_accounting}

At each step \(t\), the firm accrues expenses and revenue, which are aggregated by the controlling component into cumulative metrics. Let \(\mathrm{Cost}^{\mathrm{fix}}_t = \mathrm{fixed\_cost}\cdot dt\) denote step fixed cost, and let \(\mathrm{Cost}^c_t\) denote consultant costs from Eq.~(1). Then total step expenses are
\begin{equation}
\mathrm{Exp}_t \;=\; \mathrm{Cost}^{\mathrm{fix}}_t + \mathrm{Cost}^c_t.
\end{equation}
Cumulative revenue and expenses are computed as running sums:
\begin{equation}
\mathrm{AcRev}_T = \sum_{t=1}^{T}\mathrm{Rev}_t,
\end{equation}
\begin{equation}
\mathrm{AcExp}_T = \sum_{t=1}^{T}\mathrm{Exp}_t.
\end{equation}
Cumulative earnings (profit) are then
\begin{equation}
\mathrm{Earnings}_T \;=\; \mathrm{AcRev}_T - \mathrm{AcExp}_T.
\end{equation}
Utilization is computed from the fraction of consultant capacity actively engaged in delivery. Let \(\mathrm{busy}_{t}\) denote the number of consultants assigned to any project at step \(t\). Then step utilization is
\begin{equation}
\mathrm{Util}_t \;=\; \frac{\mathrm{busy}_{t}}{N_c},
\end{equation}
with \(\mathrm{Util}_t=0\) when \(N_c=0\). The simulator reports both per-step utilization and an overall average utilization computed across steps.

\subsubsection{Evaluation protocol in the benchmark}
\label{app:edt_eval_protocol}

For each tested model, evaluation proceeds over a fixed set of template scenarios. For each episode, the model is provided with a structured description of (i) horizon, (ii) cost parameters, and (iii) project parameters. It outputs the schema \(\{C,R,P\}\) subject to strict formatting constraints. The evaluator materializes a new scenario by applying the schema to the template (disabling projects, adjusting start/deadline windows, setting \(R\), and selecting the first \(C\) consultants). The BPTK server is then launched to execute the scenario, and the evaluator reads step-wise outputs to compute and store metrics (including cumulative earnings, revenue, expenses, cash, utilization, and revenue risk). When multiple runs per scenario are enabled, the model may adapt its schema based on prior-run feedback, enabling learning-style search over scenario configurations under a fixed action space.

\subsection{Dataset Statistics}
\label{sec:dataset_stats}

We provide the detailed sample distribution of the EnterpriseBench task corpus. Table~\ref{tab:dataset_stats} summarizes the number of samples in the training, validation, and testing sets for each component. Note that for our currently implemented \textit{online} evaluation mode, only the \textit{test} set is utilized for evaluation and sequential adaptation. 

In addition, we introduce the Beer Game and Enterprise Digital Twin (EDT) as simulation-based serious game tasks, which differ fundamentally from the sample-based datasets.

\begin{table}[h]
\centering
\small
\begin{tabular}{l|ccc|c}
\hline
\textbf{Dataset} & \textbf{Train} & \textbf{Valid} & \textbf{Test} & \textbf{Total} \\ \hline
\textit{Structured Reasoning} & & & & \\
CodeFinQA & 4,409 & 200 & 788 & 5,397 \\
CodeTAT-QA & 2,654 & 200 & 288 & 3,142 \\
ConvFinQA & 133 & - & 132 & 265 \\
FinCode & 7 & 2 & 47 & 56 \\
finer & 1,000 & 500 & 441 & 1,941 \\
FinKnow & 100 & 50 & 589 & 739 \\
formula & 500 & 300 & 200 & 1,000 \\
FormulaEval & 50 & - & 50 & 100 \\
SEC-NUM & 6,646 & 200 & 2,000 & 8,846 \\
TAT-QA & 120 & - & 120 & 240 \\ \hline
\textit{Interactive Consulting} & 12 & - & 411 & 423 \\ \hline
\textbf{Total} & \textbf{15,631} & \textbf{1,452} & \textbf{4,706} & \textbf{21,789} \\ \hline
\end{tabular}
\caption{Statistics of EnterpriseBench tasks across training, validation, and testing splits.}
\label{tab:dataset_stats}
\end{table}

\subsection{Extended Related Work}
\label{sec:extended_related_work}

\paragraph{Financial benchmarks.}
Prior work on financial and business-domain evaluation has largely focused on numerical reasoning and information extraction from static data sources. Early benchmarks such as FinQA \cite{chen2021finqa} and DocFinQA \cite{reddy2024docfinqa} target arithmetic reasoning over tables and long financial documents. Subsequent efforts, including BizBench \cite{krumdick2024bizbench}, Sec-QA \cite{lai2025sec}, and CFLUE \cite{zhu2024benchmarking}, expand task coverage across quantitative analysis and multilingual financial understanding, while FinLLMs \cite{yuan2024finllms} explores scalable benchmark construction via automatic generation. More recent holistic benchmarks, such as FinBen \cite{xie2024finben} and XFINBENCH \cite{zhang2025xfinbench}, provide broader task coverage, complemented by specialized datasets like FinTagging \cite{wang2025fintagging} and FinChain \cite{xie2026finchain} for fine-grained information structuring and reasoning verification. However, these benchmarks predominantly adopt a static QA paradigm, limiting their ability to assess interactive, sequential, and strategic decision-making processes that characterize real-world enterprise environments.

\paragraph{Reasoning and decision-making benchmarks.}
A wide variety of benchmarks have been proposed to evaluate reasoning and decision-making capabilities of agentic systems. Some focus on contextual and multi-hop reasoning, requiring models to integrate information across long contexts \cite{kuratov2024babilong, yang2018hotpotqa}. Others emphasize planning and decision-making through interaction with environments, including web-based settings \cite{miyai2025webchorearena, zhou2024webarena, deng2023mind2web, tian2025mmina, yao2022webshop} and embodied or world-like simulations \cite{shridhar2020alfworld, wang2022scienceworld, chevalier2018babyai}. Reasoning benchmarks have also been explored in complex domains such as research-oriented tasks \cite{mialon2024gaia, chen2025xbench} and tool-use scenarios \cite{qin2024toolllm}. Despite their diversity, these benchmarks are generally not designed for financial or enterprise strategic decision-making, where agents must combine professional domain knowledge, proactive information acquisition, and long-horizon trade-off reasoning.

Our proposed \textbf{EnterpriseBench} addresses these limitations by incorporating management consulting cases and serious games beyond standard QA, requiring multi-step strategic planning and cross-functional decision-making. By emphasizing interactive execution over static comprehension, EnterpriseBench offers a more realistic evaluation of agent readiness for enterprise deployment.

\subsection{Additional Backbone Results}
\label{app:additional_backbones}

To examine whether the main findings persist beyond the two backbones reported in the main text, we additionally evaluate EnterpriseBench with DeepSeek-V4-Pro and GLM-5.2. For these two additional backbones, we evaluate seven representative methods: CoT, Self-Refine, Reflexion, AMEM, Debate, Discussion, and GEPA. Tables~\ref{tab:deepseek_v4_results} and~\ref{tab:glm52_results} summarize task-level performance, while Table~\ref{tab:additional_consulting_dimensions} reports the dimension-level Consulting results.

\begin{table*}[t]
\centering
\small
\setlength{\tabcolsep}{4pt}
\begin{tabular}{ccccccccc}
\toprule
\multirow{2}{*}{Domain}
& \multirow{2}{*}{Task}
& \multicolumn{4}{c}{Single-agent}
& \multicolumn{3}{c}{Multi-agent} \\
\cmidrule(lr){3-6} \cmidrule(lr){7-9}
&
& CoT & Self-refine & Reflexion & AMEM
& Debate & Discussion & GEPA \\
\midrule

\multirow{1}{*}{Information Extraction}
& All $\uparrow$ & 0.874 & 0.876 & 0.856 & 0.864 & 0.871 & 0.895 & \best{0.939} \\
\midrule
\multirow{3}{*}{Numerical Calculation}
& Easy $\uparrow$ & 0.872 & 0.865 & 0.856 & 0.875 & 0.860 & 0.865 & \best{0.882} \\
& Middle $\uparrow$ & 0.742 & 0.776 & 0.755 & 0.776 & 0.761 & 0.776 & \best{0.784} \\
& Hard $\uparrow$ & 0.500 & \best{0.625} & 0.438 & 0.563 & 0.500 & \best{0.625} & 0.571 \\
\midrule
\multirow{3}{*}{Domain Knowledge}
& Easy $\uparrow$ & 0.962 & 0.962 & \best{0.981} & 0.943 & 0.962 & 0.962 & 0.962 \\
& Middle $\uparrow$ & 0.888 & 0.906 & 0.886 & 0.889 & 0.895 & \best{0.911} & 0.882 \\
& Hard $\uparrow$ & 0.509 & 0.519 & 0.517 & 0.472 & 0.532 & \best{0.542} & 0.495 \\
\midrule
\multirow{1}{*}{Complex Reasoning}
& All $\uparrow$ & 0.727 & 0.788 & 0.758 & 0.727 & 0.758 & \best{0.849} & 0.781 \\
\midrule
\multirow{3}{*}{\parbox{3cm}{\centering Interactive\\Decision-making}}
& Consulting $\uparrow$ & 7.06 & 6.33 & 6.63 & 7.58 & 6.29 & 6.74 & \best{8.10} \\
& BeerGame $\downarrow$ & 4.67 & 4.64 & 4.67 & \best{2.99} & 4.67 & 4.67 & 4.67 \\
& EDT $\uparrow$ & \best{6.65} & 5.66 & 5.54 & 5.54 & 5.91 & 5.69 & 5.54 \\
\bottomrule
\end{tabular}
\caption{Performance comparison of single-agent and multi-agent methods across different domains and tasks using DeepSeek-V4-Pro as the backbone model. BeerGame results are reported as total costs in units of $\times 10^4$ (lower is better), while EDT results are reported in units of $10^6$ (higher is better).}
\label{tab:deepseek_v4_results}
\end{table*}

\begin{table*}[t]
\centering
\small
\setlength{\tabcolsep}{4pt}
\begin{tabular}{ccccccccc}
\toprule
\multirow{2}{*}{Domain}
& \multirow{2}{*}{Task}
& \multicolumn{4}{c}{Single-agent}
& \multicolumn{3}{c}{Multi-agent} \\
\cmidrule(lr){3-6} \cmidrule(lr){7-9}
&
& CoT & Self-refine & Reflexion & AMEM
& Debate & Discussion & GEPA \\
\midrule

\multirow{1}{*}{Information Extraction}
& All $\uparrow$ & 0.903 & 0.898 & 0.875 & 0.907 & 0.897 & 0.911 & \best{0.932} \\
\midrule
\multirow{3}{*}{Numerical Calculation}
& Easy $\uparrow$ & 0.826 & 0.821 & 0.701 & 0.842 & 0.841 & \best{0.845} & 0.843 \\
& Middle $\uparrow$ & 0.730 & 0.746 & 0.642 & \best{0.755} & \best{0.755} & 0.746 & 0.752 \\
& Hard $\uparrow$ & 0.500 & \best{0.563} & 0.438 & \best{0.563} & \best{0.563} & \best{0.563} & \best{0.563} \\
\midrule
\multirow{3}{*}{Domain Knowledge}
& Easy $\uparrow$ & 0.925 & \best{0.962} & 0.925 & 0.943 & 0.925 & 0.906 & 0.943 \\
& Middle $\uparrow$ & \best{0.911} & 0.895 & 0.891 & 0.907 & \best{0.911} & 0.904 & 0.904 \\
& Hard $\uparrow$ & 0.522 & 0.497 & 0.501 & 0.513 & 0.517 & \best{0.534} & 0.504 \\
\midrule
\multirow{1}{*}{Complex Reasoning}
& All $\uparrow$ & 0.697 & 0.758 & 0.758 & 0.727 & 0.727 & 0.758 & \best{0.773} \\
\midrule
\multirow{3}{*}{\parbox{3cm}{\centering Interactive\\Decision-making}}
& Consulting $\uparrow$ & 7.20 & 6.46 & 6.41 & \best{7.68} & 6.81 & 6.89 & 7.57 \\
& BeerGame $\downarrow$ & 2.12 & 4.67 & 2.54 & 2.50 & 3.28 & 3.69 & \best{1.99} \\
& EDT $\uparrow$ & 7.65 & 7.08 & 7.69 & \best{7.70} & 5.22 & 6.24 & 7.65 \\
\bottomrule
\end{tabular}
\caption{Performance comparison of single-agent and multi-agent methods across different domains and tasks using GLM-5.2 as the backbone model. BeerGame results are reported as total costs in units of $\times 10^4$ (lower is better), while EDT results are reported in units of $10^6$ (higher is better).}
\label{tab:glm52_results}
\end{table*}

\begin{table*}[t]
\centering
\small
\setlength{\tabcolsep}{4.5pt}
\begin{tabular}{llccccccc}
\toprule
\multirow{2}{*}{Backbone}
& \multirow{2}{*}{Dimension}
& \multicolumn{4}{c}{Single-agent}
& \multicolumn{3}{c}{Multi-agent} \\
\cmidrule(lr){3-6} \cmidrule(lr){7-9}
& & CoT & Self-Refine & Reflexion & AMEM
& Debate & Discussion & GEPA \\
\midrule

\multirow{5}{*}{DeepSeek-V4-Pro}
& Structure
& 7.21 & 6.54 & 6.88 & 7.59 & 6.48 & 7.00 & \best{7.92} \\
& Quant.
& 6.53 & 5.50 & 5.82 & 7.20 & 5.51 & 6.03 & \best{7.68} \\
& Business
& 7.22 & 6.54 & 6.78 & 7.68 & 6.55 & 6.92 & \best{8.21} \\
& Comm.
& 7.51 & 6.95 & 7.21 & 8.15 & 6.78 & 7.19 & \best{8.51} \\
& Overall
& 7.06 & 6.33 & 6.63 & 7.58 & 6.29 & 6.74 & \best{8.10} \\

\midrule

\multirow{5}{*}{GLM-5.2}
& Structure
& 7.35 & 6.69 & 6.58 & \best{7.73} & 6.97 & 7.13 & 7.45 \\
& Quant.
& 6.71 & 5.63 & 5.60 & \best{7.25} & 6.04 & 6.20 & 7.13 \\
& Business
& 7.37 & 6.66 & 6.64 & \best{7.78} & 7.05 & 7.11 & 7.71 \\
& Comm.
& 7.67 & 7.06 & 6.97 & \best{8.27} & 7.31 & 7.34 & 7.99 \\
& Overall
& 7.20 & 6.46 & 6.41 & \best{7.68} & 6.81 & 6.89 & 7.57 \\

\bottomrule
\end{tabular}
\caption{Dimension-level performance on the Consulting task using DeepSeek-V4-Pro and GLM-5.2 as additional backbone models.}
\label{tab:additional_consulting_dimensions}
\end{table*}

The additional backbone results support the main conclusions of the paper. On Consulting, GEPA achieves the highest overall score under DeepSeek-V4-Pro and leads across all five evaluation dimensions, whereas AMEM performs best under GLM-5.2 across all five dimensions. This reversal further demonstrates that the effectiveness of an agent method depends on the underlying backbone. The foundational QA layer also continues to exhibit clear capability and difficulty effects, especially the performance drops from easy to hard numerical-calculation and domain-knowledge tasks. Moreover, stronger foundational performance does not uniformly transfer to interactive decision-making. Overall, the best-performing agent remains task- and backbone-dependent rather than being dominated by a single method.

\subsection{Expanded Human Validation of the Consulting Judge}
\label{app:expanded_judge_validation}

To complement the aggregate human audit reported in the main text, we conduct an expanded transcript-level validation of the LLM-based Consulting evaluation. We sample 100 Consulting cases and evaluate the outputs of five representative agent methods: CoT, AMEM, Self-Refine, Reflexion, and GEPA. This produces 500 method--case transcripts. Each transcript is independently scored by three human evaluators using the same information available to the LLM judge: the original case text, the complete dialogue transcript, and the four-dimensional evaluation rubric. The average of the three human ratings is used as the human reference score.

We measure transcript-level agreement using Pearson correlation, Spearman rank correlation, mean absolute error (MAE), and within-one-point agreement. Within-one-point agreement is the percentage of transcripts for which the absolute difference between the LLM score and the averaged human score is no greater than one point on the 0--10 scale. We assess inter-annotator reliability using ICC$(2,3)$, a two-way random-effects, absolute-agreement model for the average ratings of three evaluators.

\begin{table*}[t]
\centering
\small
\setlength{\tabcolsep}{5pt}
\begin{tabular}{lccccc}
\toprule
Dimension
& Pearson $r$ $\uparrow$
& Spearman $\rho$ $\uparrow$
& MAE $\downarrow$
& Within-1 $\uparrow$
& Human ICC$(2,3)$ $\uparrow$ \\
\midrule
Structure
& 0.873 & 0.743 & 0.752 & 90.4\% & 0.861 \\
Quantitative Reasoning
& 0.840 & 0.797 & 0.929 & 67.4\% & 0.782 \\
Business Sense
& 0.864 & 0.745 & 0.712 & 84.4\% & 0.834 \\
Communication
& 0.899 & 0.656 & 0.900 & 72.0\% & 0.876 \\
Overall
& 0.887 & 0.736 & 0.679 & 85.0\% & 0.892 \\
\bottomrule
\end{tabular}
\caption{Transcript-level agreement between the LLM judge and human evaluators on 500 Consulting transcripts. MAE denotes mean absolute error, Within-1 denotes agreement within one point on the 0--10 scale, and ICC$(2,3)$ measures the absolute agreement of the averaged ratings from three human evaluators.}
\label{tab:expanded_judge_validation}
\end{table*}

The expanded evaluation shows strong transcript-level alignment between the LLM judge and the human reference. Across the four rubric dimensions, Pearson correlations range from 0.840 to 0.899, while the overall score achieves a Pearson correlation of 0.887 and a Spearman correlation of 0.736. The overall MAE is 0.679, and 85.0\% of the overall scores differ from the averaged human rating by no more than one point. Human ratings also exhibit good inter-annotator reliability, with ICC$(2,3)$ values ranging from 0.782 to 0.892. Agreement is comparatively lower for quantitative reasoning and communication under the within-one-point metric, suggesting that these dimensions may benefit from stricter judge calibration.

\subsection{Resource and Adaptation Budgets}
\label{app:resource_budgets}

We measure resource usage on a shared set of 50 evaluation instances, recording all model calls and tokens, including adaptation overhead. Each method follows its original implementation and recommended configuration, with only the task interface adapted to EnterpriseBench. 

\begin{table*}[t]
\centering
\small
\setlength{\tabcolsep}{3pt}
\renewcommand{\arraystretch}{1.12}
\begin{tabular}{lcccccp{4.2cm}p{3.7cm}}
\toprule
Method
& Acc.
& Calls
& \shortstack{Calls/\\sample}
& \shortstack{Total\\tokens}
& \shortstack{Tokens/\\sample}
& \shortstack{Memory/history\\access}
& \shortstack{Prompt\\evolution} \\
\midrule
CoT
& 0.98 & 50 & 1.00 & 54,292 & 1,086
& None
& None \\

AMEM
& 0.98 & 50 & 1.00 & 52,502 & 1,050
& Online memory from previous samples only
& None \\

Self-Refine
& 0.92 & 103 & 2.06 & 130,014 & 2,600
& Current-sample trajectory only
& None \\

Reflexion
& 0.98 & 102 & 2.04 & 126,558 & 2,531
& Current-sample trajectory only
& None \\

Debate
& 0.92 & 150 & 3.00 & 206,305 & 4,126
& Current-sample discussion only
& None \\

Discussion
& 0.94 & 200 & 4.00 & 228,917 & 4,578
& Current-sample discussion only
& None \\

DC
& 0.94 & 200 & 4.00 & 972,560 & 19,451
& Online cheatsheet from previous samples only
& Online cheatsheet update \\

ACE
& 0.92 & 306 & 6.12 & 1,070,081 & 21,402
& Online experience from previous samples only
& Reflection and curation \\

GEPA
& 0.94 & 692 & 13.84 & 2,443,253 & 48,865
& Online mini-batch feedback and history only
& Prompt proposal, reflection, and merging \\
\bottomrule
\end{tabular}
\caption{Resource and adaptation budgets measured on the same 50 evaluation instances. Calls and tokens include adaptation overhead.}
\label{tab:resource_budgets}
\end{table*}

Resource consumption varies substantially across methods. CoT and AMEM achieve an accuracy of 0.98 with approximately one call and 1.1K tokens per sample. In comparison, ACE and GEPA use 6.12 and 13.84 calls per sample and approximately 21.4K and 48.9K tokens per sample, respectively, without achieving higher accuracy on this sample. These results show that a larger adaptation budget does not necessarily produce better performance. 

\subsection {AI Assistance Disclosure}

AI-based tools were used during the preparation of this work to assist with code development and to correct grammatical and stylistic issues in the manuscript. All scientific content, experimental design, results, and conclusions were conceived, implemented, and verified by the authors.

\end{document}